\documentclass[letterpaper,10 pt,conference]{ieeeconf}
\IEEEoverridecommandlockouts
\usepackage{ieee_robotics}
\usepackage{dblfloatfix}

\def\model{\texttt{Ag2x2}\xspace}

\acrodef{rl}[RL]{Reinforcement Learning}

\title{\LARGE \bf \model: Robust Agent-Agnostic Visual Representations for\\Zero-Shot Bimanual Manipulation\vspace{-12pt}}%

\author{
    Ziyin Xiong$^{1,2,3,4,5,\star}$, Yinghan Chen$^{1,2,3,4,5,6,\star}$, Puhao Li$^{1}$, Yixin Zhu$^{2,3,4,5,7}$, Tengyu Liu$^{1,\dagger}$, Siyuan Huang$^{1,\dagger}$\\
    \href{https://ziyin-xiong.github.io/ag2x2.github.io/}{https://ziyin-xiong.github.io/ag2x2.github.io/}%
    \thanks{$\star$ Ziyin Xiong and Yinghan Chen contributed equally to this work.}%
    \thanks{$\dagger$ Corresponding emails: \texttt{\{liutengyu,syhuang\}@bigai.ai}.}%
    \thanks{$^1$ National Key Laboratory of General Artificial Intelligence, Beijing Institute for General Artificial Intelligence (BIGAI).
    $^2$ School of Psychological and Cognitive Sciences, Peking University.
    $^3$ Institute for Artificial Intelligence, Peking University.
    $^4$ Beijing Key Laboratory of Behavior and Mental Health, Peking University.
    $^5$ Yuanpei College, Peking University.
    $^6$ Department of Computer Science and Technology, University of Cambridge.
    $^7$ Embodied Intelligence Lab, PKU-Wuhan Institute for Artificial Intelligence.
    This work is supported in part by the National Science and Technology Major Project (2022ZD0114900), the National Natural Science Foundation of China (62376031), the Beijing Nova Program, the State Key Lab of General AI at Peking University, the PKU-BingJi Joint Laboratory for Artificial Intelligence, and the National Comprehensive Experimental Base for Governance of Intelligent Society, Wuhan East Lake High-Tech Development Zone.}%
}

\begin{document}

\let\oldtwocolumn\twocolumn
\renewcommand\twocolumn[1][]{%
    \oldtwocolumn[{#1}{
        \centering
        \vspace{-12pt}
        \includegraphics[width=\linewidth]{teaser} 
        \captionof{figure}{\textbf{\model enables zero-shot acquisition of bimanual manipulation skills without relying on expert demonstrations or engineered rewards.} The framework operates in two key stages: (left) learning coordination-aware visual representations directly from human manipulation videos (shown in sequential frames of cooking with highlighted hand) while preserving critical hand position data despite domain differences; and (right) leveraging these representations to acquire diverse bimanual manipulation skills in simulation autonomously, demonstrated through multiple Franka robot arms performing sequential steps of various tasks including cabinet opening (top row), door manipulation (middle row), and rope handling (bottom row).}
        \label{fig:teaser}
        \vspace{6pt}
    }]
}

\maketitle
\thispagestyle{empty}
\pagestyle{empty}

\begin{abstract}
Bimanual manipulation, fundamental to human daily activities, remains a challenging task due to its inherent complexity of coordinated control. Recent advances have enabled zero-shot learning of single-arm manipulation skills through agent-agnostic visual representations derived from human videos; however, these methods overlook crucial agent-specific information necessary for bimanual coordination, such as end-effector positions. We propose \model, a computational framework for bimanual manipulation through coordination-aware visual representations that jointly encode object states and hand motion patterns while maintaining agent-agnosticism. Extensive experiments demonstrate that \model achieves a 73.5\% success rate across 13 diverse bimanual tasks from Bi-DexHands and PerAct$^2$, including challenging scenarios with deformable objects like ropes. This performance outperforms baseline methods and even surpasses the success rate of policies trained with expert-engineered rewards. Furthermore, we show that representations learned through \model can be effectively leveraged for imitation learning, establishing a scalable pipeline for skill acquisition without expert supervision. By maintaining robust performance across diverse tasks without human demonstrations or engineered rewards, \model represents a step toward scalable learning of complex bimanual robotic skills.
\end{abstract}

\section{Introduction}

Robotic manipulation today struggles with a fundamental skill that humans take for granted---using our two hands together. While we naturally coordinate our hands to cook, fold laundry, or tie shoelaces, autonomous acquisition of such bimanual manipulation skills remains a major challenge in robotics. This challenge primarily stems from two critical limitations: the dependence on human expertise for supervision and achieving precise bimanual coordination.

Current approaches to bimanual manipulation create a fundamental bottleneck in scaling robotic capabilities due to their reliance on expert input. \ac{rl} methods such as Bi-DexHands~\cite{chen2024bi} demonstrate impressive dexterity but depend on carefully engineered, task-specific rewards that rarely generalize across different tasks or embodiments. Similarly, while imitation learning approaches~\cite{kroemer2021review,li2023human,li2025controlvla} have shown promise when learning from teleoperation~\cite{zhao2023learning,fu2024mobile} and visual demonstrations~\cite{grannen2023stabilize,gao2024bi,li2025maniptrans}, collecting demonstrations for bimanual tasks is exceptionally time-consuming and resource-intensive, limiting scalability.

\begin{figure*}[t!]
    \centering
    \includegraphics[width=.95\linewidth]{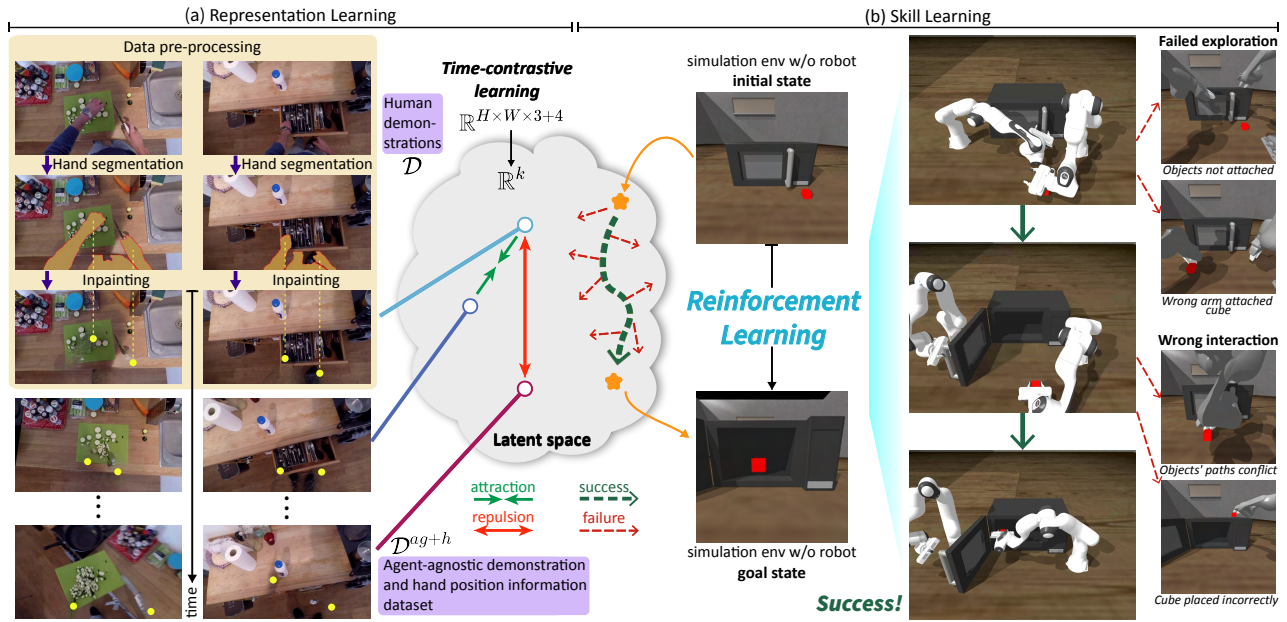} 
    \caption{\textbf{Framework of \model.} Our approach consists of two main components: (a) Representation Learning via time-contrastive learning on agent-agnostic human demonstrations with preserved hand position information, visualized through sequential cooking frames with erased hands but highlighted hand positions; and (b) Skill Learning through \ac{rl} with agent-agnostic action representations, illustrated by a robot learning to place a cube into a microwave, progressing from failed attempts to successful execution.}
    \label{fig:model}
\end{figure*}

Recent advances in single-arm manipulation~\cite{ma2022vip,nair2022r3m,li2024ag2manip} offer valuable insights for addressing these challenges by demonstrating that time-contrastive visual representations learned from existing unstructured human videos can effectively guide robotic policy learning. Human videos provide rich task-completion insights that align well with robotic objectives, making them a feasible training source. However, enabling robots to learn directly from human-centric data requires bridging the domain gap between human and robot embodiments. While some approaches~\cite{ma2022vip,nair2022r3m} struggle with bias from human kinematic structures, Ag2Manip~\cite{li2024ag2manip} attempted to solve this by masking regions occupied by humans, focusing exclusively on encoding object movements. This approach, however, eliminates all embodiment information, making it inherently unsuitable for bimanual tasks where spatial and temporal hand coordination is critical---such as transferring objects into initially closed containers. Extending these methods to bimanual manipulation thus necessitates a representation that captures both object progression and the crucial patterns of hand coordination.

To address these limitations, we present \model, a scalable framework that advances bimanual manipulation skill acquisition through coordination-aware visual representations that jointly encode object states and hand motion patterns while maintaining agent-agnosticism. Through extensive experiments across 13 diverse tasks from Bi-DexHands~\cite{chen2024bi} and PerAct$^2$~\cite{grotz2024peract2}, \model achieves a 73.5\% success rate, significantly outperforming baseline autonomous methods and even surpassing policies trained with expert-engineered rewards. Particularly noteworthy is \model's proficiency in traditionally challenging scenarios for robotics, such as manipulating deformable objects like ropes. Furthermore, we demonstrate that the skills acquired through \model can serve as high-quality training data for imitation learning, establishing an efficient pipeline for scalable robotic skill acquisition without expert supervision. By autonomously generating large-scale manipulation data while maintaining robust performance across diverse tasks, \model represents a significant step toward reducing reliance on costly human demonstrations and carefully engineered rewards, thus enabling more scalable approaches to robotic learning.

\section{Related Works}

\paragraph*{Bimanual Manipulation}

Bimanual manipulation represents a fundamental leap in complexity beyond single-arm tasks, requiring sophisticated coordination and planning capabilities~\cite{smith2012dual,mirrazavi2016coordinated,zhao2023dual,li2025maniptrans}. Current research approaches can be categorized into three distinct paradigms, each with inherent limitations for scalable deployment. 

Demonstration-based learning approaches have yielded promising results but face significant scaling barriers. While behavior cloning methods~\cite{zhao2023learning,fu2024mobile} offer framework flexibility across diverse scenarios, and structured representations like those in Bi-KVIL~\cite{gao2024bi} and PerAct$^2$~\cite{grotz2024peract2} enhance generalization, they all share a common bottleneck: dependence on extensive expert demonstrations. This requirement creates substantial data collection costs that limit real-world applicability. Even systems like 2HandedAfforder~\cite{heidinger2handedafforder}, which extract bimanual affordances from egocentric human videos, remain constrained to passive extraction without developing coordination-aware policies.

\ac{rl} approaches~\cite{chen2024bi,lin2023bi} circumvent the need for demonstrations but introduce a different yet equally problematic dependency: task-specific reward engineering. These carefully designed rewards require significant expert knowledge to develop and frequently fail to generalize beyond their target scenarios, creating another scalability barrier. 

Hybrid methods~\cite{drolet2024comparison,xie2020deep,bahety2024screwmimic} attempt to address limitations of both paradigms but typically remain restricted to narrow task domains (\eg, screw manipulation) or still require substantial expert input. This persistent reliance on human expertise---whether for demonstrations, reward engineering, or task specification---represents the critical obstacle preventing bimanual manipulation from achieving true scalability and generalization across diverse tasks.

\paragraph*{Representation Learning for Manipulation}

Visual representation learning for robotics has evolved along two complementary trajectories, each with distinct advantages and limitations for bimanual applications. 

Task-specific approaches leverage techniques such as contrastive learning~\cite{laskin2020curl} and state prediction~\cite{gelada2019deepmdp} to extract targeted features. While effective in structured environments with clear objectives, these methods struggle with cross-task generalization and typically demand substantial task-specific training data. Related techniques focusing on policy-representation decoupling~\cite{pari2021surprising} or robotic priors~\cite{jonschkowski2015learning} enhance performance in narrowly defined tasks but face similar generalization challenges across diverse manipulation scenarios.

General-purpose methods offer broader applicability through pre-trained visual representations~\cite{chen2020learning} and frameworks like RRL~\cite{shah2021rrl}. These approaches have demonstrated impressive transferability in single-arm contexts, with recent works~\cite{parisi2022unsurprising,seo2022reinforcement} validating their effectiveness. Human video-based representations like R3M~\cite{nair2022r3m}, VIP~\cite{ma2022vip}, and Ag2Manip~\cite{li2024ag2manip} further leverage everyday human activities to learn generalizable manipulation features. However, when applied to bimanual tasks, these methods reveal fundamental limitations: they either focus exclusively on single-arm scenarios, treat hands as independent agents without modeling their coordination, or fail to capture the crucial spatial relationships necessary for effective bimanual manipulation.

Our work addresses this critical gap by introducing a coordination-aware representation framework specifically engineered for bimanual tasks. Unlike existing approaches that either demand task-specific data or neglect coordination information, our method simultaneously captures object state dynamics and hand-hand interactions while maintaining agent-agnosticism. This representation enables autonomous skill acquisition without expert demonstrations or engineered rewards, while preserving the essential spatial and temporal relationships that underpin successful bimanual coordination.

\section{Method}

Our framework, \model, enables learning of bimanual manipulation skills without expert supervision. As illustrated in \cref{fig:model}, \model operates through two core components: (i) a novel representation learning approach that preserves hand coordination information from human videos, and (ii) a reward-shaping mechanism that leverages these representations to guide autonomous skill acquisition through \ac{rl}.

\subsection{From Single-Arm to Bimanual Representations}

Bimanual manipulation fundamentally differs from single-arm tasks by requiring coordinated action between two effectors. While recent work has made progress in learning manipulation from human videos, these approaches face significant limitations when extended to bimanual scenarios.

The agent-agnostic approach introduced in Li \etal~\cite{li2024ag2manip} demonstrated that removing human presence from demonstration videos could facilitate learning transferable manipulation skills. This method processes human demonstration videos $\mathcal{D}=\left\{v^{c}:=\left(o_{1}^{c}, o_{2}^{c}, \ldots, o_{n_{c}}^{c}\right)\right\}_{c=1}^{N}$ by completely masking human regions to create agent-agnostic data $\mathcal{D}^{ag}$. A visual encoder $\mathcal{F}_{\phi}: \mathbb{R}^{H \times W \times 3} \rightarrow \mathbb{R}^{K}$ then learns time-contrastive embeddings through:
\begin{equation}
    \mathcal{L}= \mathbb{E}_{o_{i}^{c}, o_{j}^{c}, o_{k}^{c}, o_{l}^{\neq c} \sim \mathcal{D}^{ag}} \mathcal{L}_{\mathrm{tcn}}+ \mathbb{E}_{o \sim \mathcal{D}^{ag}} \mathcal{L}_{\mathrm{reg}}.
    \label{eq:fullloss}
\end{equation}

This objective combines temporal contrastive learning $\mathcal{L}_{tcn}=-\log \frac{e^{S(z_i^c,z_j^c)}}{e^{S(z_i^c,z_j^c)}+e^{S(z_i^c,z_k^c)}+e^{S(z_i^c,z_l^{\neq c})}}$ with regularization $\mathcal{L}_{reg}=||\mathcal{F}_{\phi}(o)||_1+||\mathcal{F}_{\phi}(o)||_2$ to learn transferable object-centric features.

However, this approach creates a critical gap for bimanual manipulation: by removing all human presence, it eliminates essential information about spatial relationships between hands---information that is fundamental to coordination tasks like transferring objects between hands or manipulating articulated objects with synchronized movements. While focusing solely on environmental changes was sufficient for single-arm tasks, bimanual manipulation requires understanding both object state progression and the relative positioning of the manipulation agents.

Our key insight is that effective bimanual manipulation requires preserving specific coordination information while still maintaining agent-agnosticism. Rather than completely removing human presence, \model selectively retains critical positional information about hand interactions while abstracting away human-specific kinematics and appearance details that would limit transfer to robotic systems.

\subsection{Coordination-Aware Visual Representation}

Our visual encoder $\mathcal{F}_\phi$ extends Ag2Manip's architecture to jointly encode visual information and hand positions. Specifically, $\mathcal{F}_\phi:\mathbb{R}^{H\times W\times 3 + 4}\rightarrow \mathbb{R}^K$ maps the concatenation of an RGB image and the 2D coordinates of both hands to a $K$-dimensional latent embedding, enabling explicit spatial reasoning for bimanual coordination.

The encoder training follows a two-stage approach. First, we fine-tune a ViT-Large model~\cite{ViT} pre-trained on ImageNet-21k~\cite{imagenet21k} using inpainted images as the only input, employing LoRA~\cite{lora} to maintain pre-trained knowledge while enabling efficient adaptation. Second, we incorporate hand position information by introducing specialized position tokens. We encode each hand's 2D position through a two-layer MLP with hidden dimensions [16, 32] and ReLU activation, generating position-encoded tokens that are concatenated with the image tokens. For occluded or out-of-frame hands, we utilize hand-specific learnable variables. Both stages optimize the time-contrastive objective defined in \cref{eq:fullloss}, applied to our enhanced dataset $\mathcal{D}^{ag+h}$.

We derive this hand-aware agent-agnostic dataset $\mathcal{D}^{ag+h}$ from EpicKitchen~\cite{epickitchen}, an extensive collection of ego-centric videos capturing household tasks. Our processing pipeline detects hand poses using the HaMeR model~\cite{hamer} modified with a YOLO-v5~\cite{yolov5} backbone for improved efficiency, projects 3D key points to 2D coordinates using estimated camera poses, computes each hand's position as the mean of its 21 key points, and generates agent-agnostic frames through ODISE~\cite{odise} segmentation followed by E$^2$FGVI~\cite{e2fgvi} inpainting. The resulting dataset comprises frames $o^c_i\in\mathbb{R}^{H\times W\times 3+4}$ that combine processed RGB images with corresponding hand coordinates.

\subsection{Policy Learning and Reward Shaping}

We extend agent-agnostic action representation to bimanual manipulation by learning coordinated motions and forces for two free-floating proxy agents, subsequently using inverse kinematics to convert these proxy trajectories into robot joint commands. For effective grasping, we first identify feasible graspable poses for target objects using GraspNet~\cite{fang2020graspnet}. Each end-effector is represented as an agent-agnostic sphere, with a grasp considered valid when the proxy's distance from the pose is less than 5cm, indicating successful attachment.

With these graspable regions established, we define our policy learning objective. Given a goal specification $g\in \mathbb{R}^{H\times W\times 3+4}$, comprising both an image of the goal state and 2D projections of desired end-effector positions in camera space, we employ model-free \ac{rl} to obtain an action policy $\pi$ that maps proxy states $p_t$ and environment states $s_t$ to actions $a_t=(a_p^t, a_f^t)$. Here, $a_p^t\in\mathbb{R}^6$ specifies target positions for both proxies during exploration, while $a_f^t\in\mathbb{R}^6$ defines their intended forces during interaction. A PD controller translates these target actions into actual proxy movements.

Our reward function is designed to maximize embedding similarity between current state $o_t$ and goal state $g$ while promoting effective exploration. Rather than penalizing deviations from a single optimal trajectory, we introduce a tilted reward formulation:
\begin{equation}
    \resizebox{.91\linewidth}{!}{$\displaystyle%
        \mathcal{R}(o_t,g;\phi) = \exp((1+\alpha\cdot \textbf{1}_{S(z_t,z_g)>\beta})\frac{S(z_t,z_g)-\beta}{\beta})-1,%
    $}%
\end{equation}
where $\beta = S(z_0,z_g)$ denotes the initial embedding similarity between start and goal states, and $\alpha>0$ controls reward scaling. This approach improves upon traditional similarity-based rewards by requiring states to exceed the initial condition rather than merely maximizing consecutive improvements, facilitating more effective exploration during early training when policy behavior is predominantly stochastic.

\subsection{Implementation Details}

Our implementation consists of two key components: the visual encoder architecture and the policy learning framework. For the visual encoder, we process images resized to $224\times 224\times 3$ dimensions and implement LoRA with rank $r=4$ for the query and value matrices in all self-attention layers. The encoder training proceeds sequentially: first, the visual adaptation phase runs for 40 hours on 4 NVIDIA A100 GPUs using Adam optimization with a learning rate of $10^{-4}$, followed by the hand-position integration phase requiring 20 hours on 8 NVIDIA A100 GPUs.

For the \ac{rl} component, we employ Proximal Policy Optimization (PPO)~\cite{schulman2017proximal} to train our bimanual manipulation policy. The proxy agents are configured with a sphere collision radius of 2cm and an interactive region radius of 5cm. Based on ablation studies, we set the reward scaling parameter $\alpha=3.0$. Each manipulation skill requires approximately 3 hours of training on a single NVIDIA 3090 workstation.

\section{Experiments}

We evaluate \model's effectiveness in bimanual manipulation through extensive experiments across diverse manipulation scenarios. Our approach achieves a 73.5\% average success rate across 13 challenging tasks requiring coordinated two-handed interaction, significantly outperforming existing autonomous methods and policies trained with expert-designed rewards. This section details our experimental setup (\cref{sec:exp:tasks}), presents quantitative performance results (\cref{sec:exp:result}), analyzes key design choices through ablation studies (\cref{sec:exp:ablation}), and provides additional experimental insights (\cref{sec:exp:additional}).

\subsection{Experimental Setup}\label{sec:exp:tasks}

\paragraph*{Task Selection}

We curated 13 diverse bimanual manipulation tasks from two established benchmarks: 6 tasks from Bi-DexHands \cite{chen2024bi} and 7 from PerAct$^2$ \cite{grotz2024peract2}. We selected tasks based on two criteria: compatibility with gripper-based manipulation and single-stage structure to ensure well-defined goal specifications. Multi-stage tasks were excluded as they require intermediate goal definitions beyond our current framework.

\paragraph*{Implementation}

All experiments use two 9-DoF Franka Emika robotic arms with integrated grippers (each with 7 arm joints and 2 gripper fingers), simulated in NVIDIA IsaacGym to leverage its GPU acceleration capabilities for efficient reinforcement learning. For Bi-DexHands tasks, we directly utilized existing object assets. For PerAct$^2$ tasks, originally implemented in CoppeliaSim, we converted assets from \textit{.ttm} to \textit{.urdf} and \textit{.dae} formats for IsaacGym compatibility. When direct asset migration was not possible, we substituted comparable alternatives to maintain task integrity.

\paragraph*{Evaluation Protocol}

Each task begins with the robotic arms in predefined positions, with goals specified by images capturing the target state from one of three camera angles (center, up, down). A task is considered successful when objects reach the goal configuration within an acceptable error margin. To ensure robust evaluation, we test each task under nine different configurations by combining three initialization seeds with three camera perspectives. All evaluated models were trained with 68 parallel environments over 200 episodes per task.

\paragraph*{Comparison to Benchmarks}

Unlike PerAct$^2$, our approach does not require expert-designed waypoints to guide end-effectors through specific intermediate targets, relying only on the final position information. Despite this reduced guidance, \model achieves significantly higher success rates on most PerAct$^2$ tasks than those reported in the original benchmark \cite{grotz2024peract2}, demonstrating its superior robustness and adaptability to complex bimanual manipulation scenarios.

\begin{figure*}[b!]
    \centering
    \includegraphics[width=\linewidth]{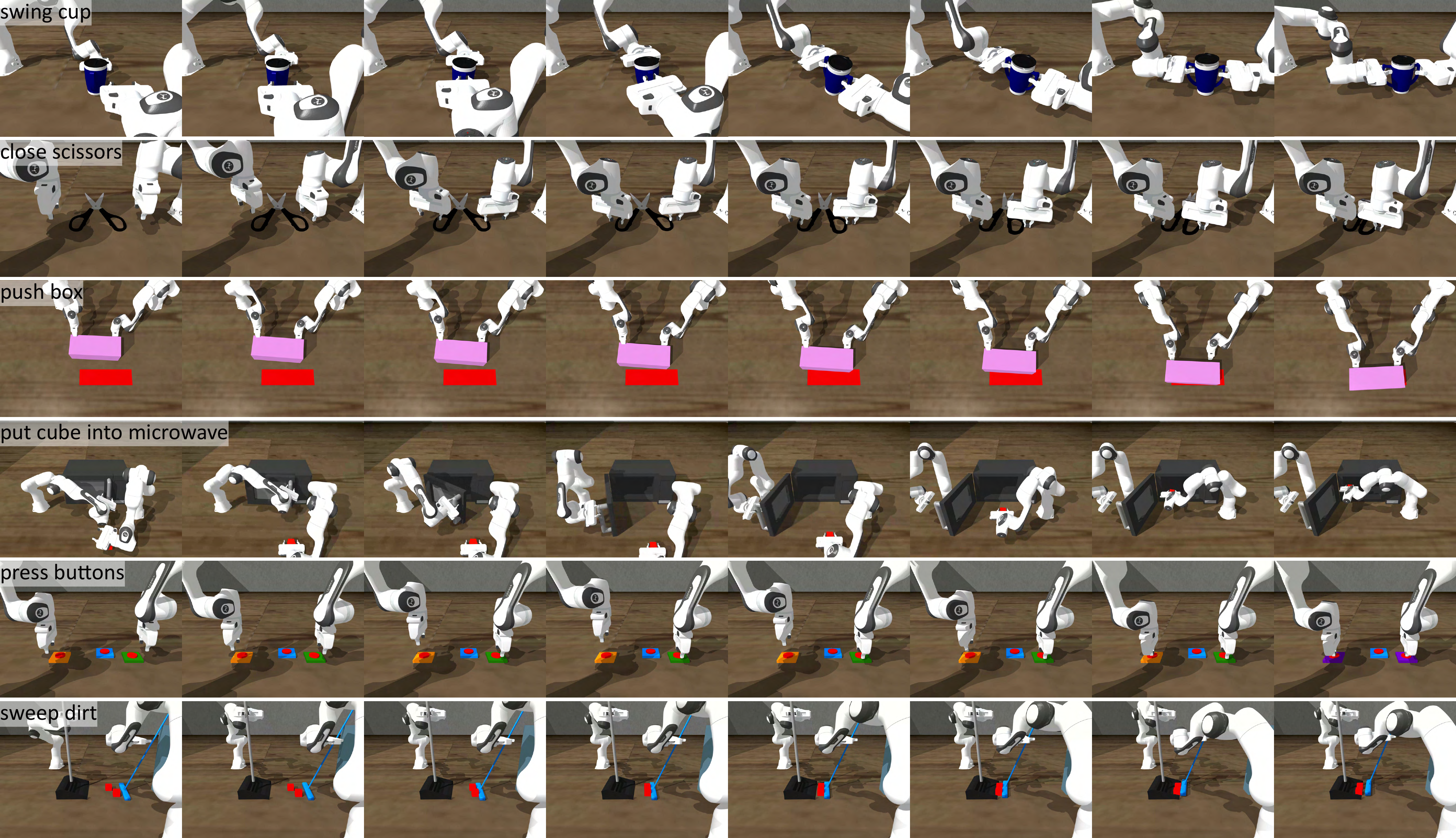}
    \caption{\textbf{Qualitative results of bimanual manipulation.} Temporal sequences showing successful task execution by our learned policies across six representative tasks: swinging a cup, closing scissors, pushing a box, placing a cube into a microwave, pressing buttons, and sweeping dirt. Each row presents the progression of a single task from left to right. Our method generates coordinated trajectories for both end-effectors, which are converted to joint commands for the two Franka arms through inverse kinematics. Note the synchronized bimanual coordination required across diverse manipulation scenarios, from precise interactions (scissors, buttons) to force-sensitive operations (pushing box, sweeping).}
    \label{fig:results}
\end{figure*}

\subsection{Experimental Results}\label{sec:exp:result}

\paragraph*{Baselines}

We benchmark \model against leading approaches in zero-shot skill learning, including agent-aware visual representations (R3M~\cite{nair2022r3m} and VIP~\cite{ma2022vip}), agent-agnostic representation (Ag2Manip~\cite{li2024ag2manip}), and LLM-driven autonomous reward generation (Eureka~\cite{ma2023eureka}). For Eureka, we disable human feedback to evaluate solely the method's intrinsic learning capabilities without task-specific expert knowledge. Our strongest baseline uses expert-designed rewards that directly minimize distances between key object poses and goal states.

\begin{table}[t!]  
	\centering
    \small
	\caption{\textbf{Comparison and ablation results.} Success rates across 13 bimanual manipulation tasks from two benchmarks. Success counts range from 0-9, representing successful attempts out of 9 evaluation runs. \textbf{Methods compared:} R3M, VIP, and Ag2Manip are visual representation approaches; Eureka uses LLM-generated rewards without human feedback; expert reward employs manually-designed reward functions minimizing distance between object poses and goal states; \model-H is our ablation that excludes hand-related features from reward computation; \model is our full model with hand feature integration. \textbf{Evaluation protocols:} Our method and its ablation, R3M, VIP, and Ag2Manip were evaluated using 3 seeds × 3 camera views; Eureka using 3 seeds × 3 different generated reward functions; and expert reward using 9 different seeds. \textbf{Tasks include:} From Bi-DexHands (a-f): close door outward, close door inward, open pen cap, lift pot, swing cup, close scissors; From PerAct$^2$ (g-m): push box, put cube into drawer, put cube into microwave, lift tray, press buttons, sweep dirt, straighten rope.}
	\label{tab:methodcompare}
    \setlength{\tabcolsep}{3pt}
    \resizebox{\linewidth}{!}{%
    	\begin{tabular}{cccccccccccccccccc}
    		\toprule
    		\multirow{2}*{Method} & \multicolumn{7}{c}{Bi-DexHands} &&  \multicolumn{8}{c}{PerAct$^2$} & \multirow{2}*{Overall} \\
    		\cline{2-8}\cline{10-17}
    		  &\bf a&\bf b&\bf c&\bf d&\bf e&\bf f&\bf Avg.&&\bf g&\bf h&\bf i&\bf j&\bf k&\bf l&\bf m&\bf Avg.&\\
            \midrule
            Eureka~\cite{ma2023eureka} &0&0&0&2&1&5&14.8\%&&0&1&0&0&7&2&0&15.9\%&15.4\%\\
            R3M~\cite{nair2022r3m} &0&0&3&0&1&0&7.4\%&&2&0&\underline{4}&2&3&\underline{3}&0&22.2\%&15.4\%\\
            VIP~\cite{ma2022vip} &1&3&1&7&2&0&25.9\%&&0&0&\underline{4}&\underline{5}&5&\underline{3}&0&27.0\%&26.5\%\\
            Ag2Manip~\cite{li2024ag2manip} &6&\bf 9&\underline{7}&4&3&\underline{7}&66.7\%&&2&3&3&3&\bf 9&\bf 6&4&47.6\%&56.4\%\\
            expert reward &\textbf{8}&\bf 9&6&6&\bf 8&\bf 9&\bf 85.2\%&&\underline{5}&0&\bf 6&3&5&\underline{3}&\bf 6&44.4\%&\underline{63.2}\%\\
            \midrule
    		\model-H  &7&4&\underline{7}&\underline{7}&4&\bf 9&\underline{70.4}\%&&\underline{5}&\underline{4}&3&\underline{5}&\underline{8}&\underline{3}&3&\underline{46.0}\%&57.3\% \\
            \model &\underline{7}&\underline{6}&\bf 9&\bf 8&\underline{7}&\bf 9&\bf 85.2\%&&\bf 6&\bf 5&2&\bf 7&\bf 9&\bf 6&\underline{5}&\bf 63.5\%&\bf 73.5\% \\
    		\bottomrule
    	\end{tabular}%
    }%
\end{table}

\paragraph*{Performance Analysis}

As shown in \cref{tab:methodcompare}, \model achieves a 73.5\% average success rate across all tasks, substantially outperforming Ag2Manip (56.4\%) and even expert-designed rewards (63.2\%). This improvement is particularly notable given that \model requires no manual reward engineering or task-specific demonstrations.

Method performance varies significantly across tasks. Agent-aware representations (R3M, VIP) consistently underperform due to the visual and kinematic gap between humans and robots. Ag2Manip, while effective for single-handed tasks, struggles with bimanual coordination in scenarios like box pushing and tray lifting. Eureka performs well only on tasks with easily describable goal states (closing scissors, pressing buttons) but falters on more complex interactions. Expert-designed rewards show strong performance on most tasks but fail in scenarios requiring temporal coordination between hands and objects, such as \textit{(h) put cubes into drawer}, \textit{(j) lift tray}, and \textit{(l) sweep dirt}.

\model demonstrates remarkable consistency across diverse manipulation challenges. It successfully learns 12 of 13 tasks, with only \textit{(i) put cube into microwave} falling below 50\% success rate. Most impressively, it masters deformable object manipulation (\textit{(m) straighten rope}) without specialized rewards or demonstrations, suggesting broader applicability to complex household tasks like laundry organization. \cref{fig:results} illustrates successful trajectories from representative tasks.

\begin{figure*}[t!]
    \centering
    \includegraphics[width=\linewidth]{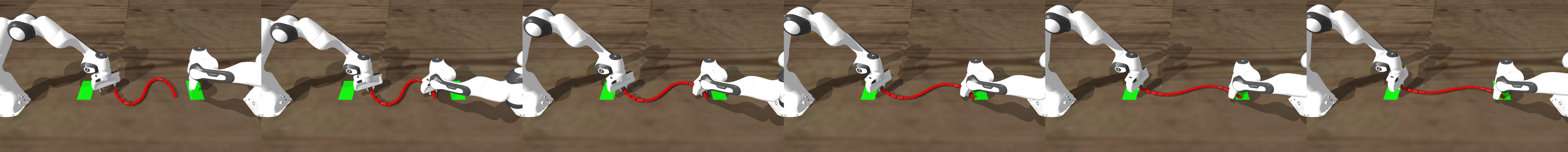}
    \caption{\textbf{Generalization of learned manipulation skills through imitation policy rollout.} This sequence demonstrates an imitation policy successfully straightening a rope from an unseen initial configuration. The policy was trained on just 12 expert trajectories generated by \model and successfully transfers to novel states. Images progress from left to right, showing coordinated bimanual manipulation that adapts to the deformable object's changing state.}
    \label{fig:imitation_learning}
\end{figure*}

\begin{table*}[t!]
    \centering
    \small
    \caption{\textbf{Trajectory smoothness analysis.} Lower values indicate smoother movements with reduced end-effector acceleration throughout task execution. All values represent cumulative acceleration magnitude of both end-effectors across successful task completions only, proportional to $m/s^2$ by a factor of 36. Dashes (-) indicate methods that failed all evaluation runs for that task. For detailed descriptions of all methods (VIP, Ag2Manip, expert reward, \model-H, and \model), refer to \cref{tab:methodcompare}. Tasks correspond to those listed in \cref{tab:methodcompare}.}
	\label{tab:traj_smooth}
    \setlength{\tabcolsep}{3pt}
    \begin{tabular}{cccccccccccccccc}
        \toprule
        \multirow{2}*{Method} & \multicolumn{6}{c}{Bi-DexHands} &&  \multicolumn{7}{c}{PerAct$^2$} & \multirow{2}*{Avg.} \\
        \cline{2-7}\cline{9-15}
          &\bf a&\bf b&\bf c&\bf d&\bf e&\bf f&&\bf g&\bf h&\bf i&\bf j&\bf k&\bf l&\bf m&\\
        \midrule
        VIP~\cite{ma2022vip} &2.38&2.11&1.24&1.97&1.85&-&&-&-&2.03&1.69&0.56&0.37&-&1.58\\
        Ag2Manip~\cite{li2024ag2manip} &1.62&1.62&1.35&1.46&1.60&0.45&&1.74&3.98&1.77&1.11&0.48&0.38&0.61&1.36\\
        expert reward &1.70&1.68&1.15&1.00&1.06&0.70&&0.58&-&1.99&1.64&0.49&0.39&0.95&\bf 1.11\\
        \model-H  &1.65&1.79&1.10&1.04&1.66&0.75&&0.67&1.37&1.50&1.43&0.59&1.30&1.13&1.23 \\
        \model &1.62&1.70&1.04&1.53&1.61&0.70&&0.48&1.13&2.00&1.24&0.54&0.36&1.04&\underline{1.15} \\
        \bottomrule
    \end{tabular}
\end{table*}

\paragraph*{Failure Case Analysis}

We identify two primary failure patterns for future improvements. First, object interference significantly impacts performance in container-based tasks like \textit{(h) put cube into drawer} and \textit{(i) put cube into microwave}. In these scenarios, the visual salience of container movements overshadows object placement and hand motion, leading to suboptimal reward and reduced accuracy. Second, imperfect bimanual coordination occasionally emerges in challenging scenarios. In \textit{(b) close door inward}, failures often occur when one hand closes the door prematurely before the other hand positions correctly. Similarly, in \textit{(m) straighten rope}, a common failure pattern involves one hand reaching the target position while the other misses by small margins.

These observations suggest future work should focus on enhancing feature selection mechanisms for bimanual coordination and developing more sophisticated attention frameworks for multi-object interactions.

\begin{table*}[t!]  
    \centering
    \small
    \caption{\textbf{Task progress consistency analysis.} Higher values indicate more direct, monotonic progress toward the goal state. Values represent Spearman Rank Correlation between temporal frame sequence and visual similarity to the final goal state, measuring how consistently the agent progresses toward task completion. Positive values near 1.0 indicate steady progress, while values near 0 or negative suggest inefficient paths with unnecessary motions or regressions. Dashes (-) indicate methods that failed all evaluation runs for that task. For detailed descriptions of all methods and tasks, refer to \cref{tab:methodcompare}.}
	\label{tab:progress_consistency}
    \setlength{\tabcolsep}{3pt}
	\begin{tabular}{cccccccccccccccc}
		\toprule
		\multirow{2}*{Method} & \multicolumn{6}{c}{Bi-DexHands} &&  \multicolumn{7}{c}{PerAct$^2$} & \multirow{2}*{Avg.} \\
		\cline{2-7}\cline{9-15}
		  &\bf a&\bf b&\bf c&\bf d&\bf e&\bf f&&\bf g&\bf h&\bf i&\bf j&\bf k&\bf l&\bf m&\\
        \midrule
        VIP~\cite{ma2022vip} &.724&.900&.021&.136&-.154&-&&-&-&.699&-.247&-.249&.056&-&.210\\
        Ag2Manip~\cite{li2024ag2manip} &.797&.851&.692&.302&.719&.398&&.942&.025&.784&.405&-.244&-.092&.974&.504\\
		\model-H &.824&.820&.789&.094&.288&.669&&.908&.106&.719&.575&-.223&.136&.981&\underline{.514} \\
        \model &.822&.812&.685&.192&.517&.804&&.920&.018&.712&.364&-.185&.077&.963&\bf .516 \\
		\bottomrule
	\end{tabular}
\end{table*}

\subsection{Ablation}\label{sec:exp:ablation}

To understand the impact of hand position information on bimanual coordination, we conducted an ablation study. In \cref{tab:methodcompare}, we present ``\model-H'' (without hands), which learns directly from videos where human hands have been inpainted without hand position tokens provided to the model.

This ablation reveals three significant findings. First, removing hand position information causes a consistent 16.2\% performance drop across our task suite, highlighting this feature's critical role in facilitating precise bimanual coordination. Second, performance across tasks remains generally consistent with one notable exception: the microwave cube placement task \textit{(i)} shows improved performance in the ablated model. This counter-intuitive result occurs because the container's state changes visually dominate this task, overshadowing both hand positioning and object motion signals in reward computation. Third, our ablated model performs similarly to Ag2Manip despite architectural differences, confirming that \model's superior performance stems primarily from the incorporation of hand positioning rather than from switching from ResNet to ViT backbone.

Despite removing hand position information, our ablated model still outperforms most baseline methods and approaches the effectiveness of expert-designed rewards. This suggests our core architectural design remains robust even without explicit hand positioning cues, while still demonstrating that hand positioning substantially enhances performance for complex bimanual tasks.

\subsection{Imitation Learning}

Using \model, we collected 12 demonstration trajectories of rope straightening across various initial configurations. From these demonstrations, we trained an imitation learning policy that successfully generalizes to novel rope configurations. \Cref{fig:imitation_learning} illustrates the learned policy performing on an unseen scenario where the rope's bending direction is reversed from training examples. This ability to quickly collect high-quality demonstration data and learn generalizable policies highlights \model's potential for scaling up bimanual manipulation data collection, addressing a critical bottleneck in robot learning.

\begin{table*}[t!]  
	\centering
    \small
	\caption{\textbf{Impact of proprioception on performance.} Success rates when explicitly incorporating proprioceptive information (end-effector positions) in policy training. Success counts range from 0-9, representing successful attempts out of 9 evaluation runs (3 seeds × 3 camera views). All three methods use the same visual backbone but differ in their use of proprioceptive signals: Ag2Manip with limited proprioception, \model-H with partial integration, and \model with our complete proprioceptive integration approach. For detailed task descriptions, refer to \cref{tab:methodcompare}.}
	\label{tab:proprioception}
    \setlength{\tabcolsep}{3pt}
	\begin{tabular}{cccccccccccccccccc}
		\toprule
		\multirow{2}*{Method} & \multicolumn{7}{c}{Bi-DexHands} &&  \multicolumn{8}{c}{PerAct$^2$} & \multirow{2}*{Overall} \\
		\cline{2-8}\cline{10-17}
		  &\bf a&\bf b&\bf c&\bf d&\bf e&\bf f&\bf Avg.&&\bf g&\bf h&\bf i&\bf j&\bf k&\bf l&\bf m&\bf Avg.&\\
        \midrule
        Ag2Manip~\cite{li2024ag2manip} &6&9&5&4&5&8&68.5\%&&2&1&2&2&9&5&5&41.3\%&53.8\%\\
        \model-H  &7&8&5&6&3&9&70.4\%&&3&0&6&3&8&4&5&46.0\%&57.3\% \\
        \model &7&9&9&7&7&9&88.9\%&&5&2&4&3&9&5&5&52.4\%&69.2\% \\
		\bottomrule
	\end{tabular}
\end{table*}

\subsection{Additional Analysis}\label{sec:exp:additional}

We examine trajectory characteristics, reward quality, and proprioception to provide deeper insights into \model's performance advantages.

\paragraph*{Trajectory smoothness}

We quantify motion efficiency through cumulative end-effector acceleration:
\begin{equation}
    \gamma_{s} = \sum_{t=2}^{T}||\alpha_t^L||_2 + ||\alpha_t^R||_2,
\end{equation}
where $\alpha_t^L, \alpha_t^R$ denote the acceleration of left and right end-effectors at time $t$. Lower $\gamma{s}$ values indicate smoother trajectories with fewer abrupt velocity changes. As shown in \cref{tab:traj_smooth}, \model generates trajectories with smoothness comparable to expert-designed rewards and significantly better than the highest-performing baseline methods, demonstrating the advantage of coordination-aware representations in generating stable end-effector motions.

\paragraph*{Reward Consistency}

To evaluate how well our visual representations align with task progress, we use Spearman Rank Correlation \cite{spearman1961proof} to measure the relationship between temporal sequence and goal-state similarity. \Cref{tab:progress_consistency} shows \model achieves superior correlation compared to baselines, indicating more consistent task progression representation.

Our analysis also reveals an important limitation of correlation metrics for certain tasks. For instance, the button-pressing task \textit{(k)} shows a negative correlation despite achieving a 100\% success rate. This occurs because discrete state changes (button color transitions) create discontinuities in the visual representation space that don't align with steady progress measurement.

\paragraph*{Proprioception}

To determine whether \model's gains stem from hand information during pre-training or proprioception (specifically, end-effector positions) during policy learning, we conducted a controlled experiment. We compared three models (Ag2Manip, \model, and \model-H), adding a reward term to explicitly include proprioception by encouraging end-effector alignment with target positions. As shown in \cref{tab:proprioception}, the performance differences between models with proprioception remain consistent with our main results in \cref{tab:methodcompare}. This confirms that \model's advantages primarily stem from incorporating hand information during pre-training rather than from proprioceptive signals during policy learning.

\section{Conclusion}

We introduced \model, a framework addressing zero-shot bimanual manipulation through coordination-aware visual representations. By simultaneously encoding environmental state and hand positions, \model achieves a 73.5\% success rate across 13 complex bimanual tasks, significantly outperforming baselines and matching expert-designed policies. Our approach demonstrates particular strength with deformable object manipulation and supports effective imitation learning, establishing a foundation for scalable robotic manipulation.

\paragraph*{Limitations}

Our method is constrained by its representation of goals as single static images with end-effector positions. Without intermediate trajectory states, \model cannot support tasks where intermediate dynamics are essential. For example, when tasked with ``throw a ball into a bin'' with only the final goal image showing the ball in the bin, the system would attempt to directly place rather than throw the ball. Similarly, tasks with unobservable dynamics (like turning on a microwave without visual feedback on its internal state) remain challenging. While a high-level planner generating intermediate visual and kinematic steps could address some issues, fundamental limitations remain.

Additionally, our current approach only models end-effector positions, limiting applications to simple grippers rather than complex hands with articulated fingers. Extending \model to dexterous manipulation would require substantial work in understanding intricate finger coordination beyond our current scope. Nevertheless, we believe our approach of using representations pre-trained on readily available human videos provides a promising foundation for improving learning efficiency even for more complex manipulation tasks.

\paragraph*{Implications}

\model enables autonomous generation of high-quality demonstrations, reducing reliance on human-generated data in robotic training. This creates opportunities for accelerating robotic learning across diverse environments. By enabling autonomous learning of complex bimanual manipulation, our work advances generalizable robot capabilities for sophisticated physical interactions. The demonstrated success suggests coordination-aware visual representations offer a promising direction for robot manipulation.

\small
\bibliographystyle{IEEEtran}
\bibliography{reference_header,reference}

\begin{thebibliography}{10}
\providecommand{\url}[1]{#1}
\csname url@samestyle\endcsname
\providecommand{\newblock}{\relax}
\providecommand{\bibinfo}[2]{#2}
\providecommand{\BIBentrySTDinterwordspacing}{\spaceskip=0pt\relax}
\providecommand{\BIBentryALTinterwordstretchfactor}{4}
\providecommand{\BIBentryALTinterwordspacing}{\spaceskip=\fontdimen2\font plus
\BIBentryALTinterwordstretchfactor\fontdimen3\font minus \fontdimen4\font\relax}
\providecommand{\BIBforeignlanguage}[2]{{%
\expandafter\ifx\csname l@#1\endcsname\relax
\typeout{** WARNING: IEEEtran.bst: No hyphenation pattern has been}%
\typeout{** loaded for the language `#1'. Using the pattern for}%
\typeout{** the default language instead.}%
\else
\language=\csname l@#1\endcsname
\fi
#2}}
\providecommand{\BIBdecl}{\relax}
\BIBdecl

\bibitem{chen2024bi}
Y.~Chen, Y.~Geng, F.~Zhong, J.~Ji, J.~Jiang, Z.~Lu, H.~Dong, and Y.~Yang, ``Bi-dexhands: Towards human-level bimanual dexterous manipulation,'' \emph{{Transactions on Pattern Analysis and Machine Intelligence (TPAMI)}}, vol.~46, no.~5, pp. 2804--2818, 2024.

\bibitem{kroemer2021review}
O.~Kroemer, S.~Niekum, and G.~Konidaris, ``A review of robot learning for manipulation: Challenges, representations, and algorithms,'' \emph{{Journal of Machine Learning Research (JMLR)}}, vol.~22, no.~30, pp. 1--82, 2021.

\bibitem{li2023human}
J.~Li, J.~Wang, S.~Wang, and C.~Yang, ``Human--robot skill transmission for mobile robot via learning by demonstration,'' \emph{Neural Computing and Applications}, vol.~35, no.~32, pp. 23\,441--23\,451, 2023.

\bibitem{li2025controlvla}
P.~Li, Y.~Wu, Z.~Xi, W.~Li, Y.~Huang, Z.~Zhang, Y.~Chen, J.~Wang, S.-C. Zhu, T.~Liu \emph{et~al.}, ``Controlvla: Few-shot object-centric adaptation for pre-trained vision-language-action models,'' \emph{arXiv preprint arXiv:2506.16211}, 2025.

\bibitem{zhao2023learning}
T.~Z. Zhao, V.~Kumar, S.~Levine, and C.~Finn, ``Learning fine-grained bimanual manipulation with low-cost hardware,'' in \emph{{Robotics: Science and Systems (RSS)}}, 2023.

\bibitem{fu2024mobile}
Z.~Fu, T.~Z. Zhao, and C.~Finn, ``Mobile aloha: Learning bimanual mobile manipulation with low-cost whole-body teleoperation,'' in \emph{{Conference on Robot Learning (CoRL)}}, 2024.

\bibitem{grannen2023stabilize}
J.~Grannen, Y.~Wu, B.~Vu, and D.~Sadigh, ``Stabilize to act: Learning to coordinate for bimanual manipulation,'' in \emph{{Conference on Robot Learning (CoRL)}}, 2023.

\bibitem{gao2024bi}
J.~Gao, X.~Jin, F.~Krebs, N.~Jaquier, and T.~Asfour, ``Bi-kvil: Keypoints-based visual imitation learning of bimanual manipulation tasks,'' in \emph{{IEEE International Conference on Robotics and Automation (ICRA)}}, 2024.

\bibitem{li2025maniptrans}
K.~Li, P.~Li, T.~Liu, Y.~Li, and S.~Huang, ``Maniptrans: Efficient dexterous bimanual manipulation transfer via residual learning,'' in \emph{{Proceedings of Conference on Computer Vision and Pattern Recognition (CVPR)}}, 2025.

\bibitem{ma2022vip}
Y.~J. Ma, S.~Sodhani, D.~Jayaraman, O.~Bastani, V.~Kumar, and A.~Zhang, ``{VIP:} towards universal visual reward and representation via value-implicit pre-training,'' in \emph{{Proceedings of International Conference on Learning Representations (ICLR)}}, 2023.

\bibitem{nair2022r3m}
S.~Nair, A.~Rajeswaran, V.~Kumar, C.~Finn, and A.~Gupta, ``{R3M:} {A} universal visual representation for robot manipulation,'' in \emph{{Conference on Robot Learning (CoRL)}}, 2022.

\bibitem{li2024ag2manip}
P.~Li, T.~Liu, Y.~Li, M.~Han, H.~Geng, S.~Wang, Y.~Zhu, S.-C. Zhu, and S.~Huang, ``Ag2manip: Learning novel manipulation skills with agent-agnostic visual and action representations,'' in \emph{{IEEE/RSJ International Conference on Intelligent Robots and Systems (IROS)}}, 2024.

\bibitem{grotz2024peract2}
M.~Grotz, M.~Shridhar, T.~Asfour, and D.~Fox, ``Peract2: Benchmarking and learning for robotic bimanual manipulation tasks,'' \emph{CoRL 2024 Workshop on Whole-Body Control and Bimanual Manipulation (CoRL 2024 WCBM)}, 2024.

\bibitem{smith2012dual}
C.~Smith, Y.~Karayiannidis, L.~Nalpantidis, X.~Gratal, P.~Qi, D.~V. Dimarogonas, and D.~Kragic, ``Dual arm manipulation—a survey,'' \emph{Robotics and Autonomous Systems}, vol.~60, no.~10, pp. 1340--1353, 2012.

\bibitem{mirrazavi2016coordinated}
S.~S. Mirrazavi~Salehian, N.~B. Figueroa~Fernandez, and A.~Billard, ``Coordinated multi-arm motion planning: Reaching for moving objects in the face of uncertainty,'' in \emph{{Robotics: Science and Systems (RSS)}}, 2016.

\bibitem{zhao2023dual}
X.~Zhao, Y.~Zhang, W.~Ding, B.~Tao, and H.~Ding, ``A dual-arm robot cooperation framework based on a nonlinear model predictive cooperative control,'' \emph{{IEEE/ASME Transactions on Mechatronics (T-MECH)}}, vol.~29, pp. 3993--4005, 2024.

\bibitem{heidinger2handedafforder}
M.~Heidinger, S.~Jauhri, V.~Prasad, and G.~Chalvatzaki, ``2handedafforder: Learning precise actionable bimanual affordances from human videos,'' in \emph{{Conference on Robot Learning (CoRL)}}, 2024.

\bibitem{lin2023bi}
Y.~Lin, A.~Church, M.~Yang, H.~Li, J.~Lloyd, D.~Zhang, and N.~F. Lepora, ``Bi-touch: Bimanual tactile manipulation with sim-to-real deep reinforcement learning,'' \emph{{IEEE Robotics and Automation Letters (RA-L)}}, vol.~8, no.~9, pp. 5472--5479, 2023.

\bibitem{drolet2024comparison}
M.~Drolet, S.~Stepputtis, S.~Kailas, A.~Jain, J.~Peters, S.~Schaal, and H.~B. Amor, ``A comparison of imitation learning algorithms for bimanual manipulation,'' \emph{{IEEE Robotics and Automation Letters (RA-L)}}, 2024.

\bibitem{xie2020deep}
F.~Xie, A.~Chowdhury, M.~C. De~Paolis~Kaluza, L.~Zhao, L.~L. Wong, and R.~Yu, ``Deep imitation learning for bimanual robotic manipulation,'' in \emph{{Proceedings of Advances in Neural Information Processing Systems (NeurIPS)}}, 2020.

\bibitem{bahety2024screwmimic}
A.~Bahety, P.~Mandikal, B.~Abbatematteo, and R.~Mart{\'\i}n-Mart{\'\i}n, ``Screwmimic: Bimanual imitation from human videos with screw space projection,'' in \emph{{Robotics: Science and Systems (RSS)}}, 2024.

\bibitem{laskin2020curl}
M.~Laskin, A.~Srinivas, and P.~Abbeel, ``Curl: Contrastive unsupervised representations for reinforcement learning,'' in \emph{{Proceedings of International Conference on Machine Learning (ICML)}}, 2020.

\bibitem{gelada2019deepmdp}
C.~Gelada, S.~Kumar, J.~Buckman, O.~Nachum, and M.~G. Bellemare, ``Deepmdp: Learning continuous latent space models for representation learning,'' in \emph{{Proceedings of International Conference on Machine Learning (ICML)}}, 2019.

\bibitem{pari2021surprising}
J.~Pari, N.~M. Shafiullah, S.~P. Arunachalam, and L.~Pinto, ``The surprising effectiveness of representation learning for visual imitation,'' in \emph{{Robotics: Science and Systems (RSS)}}, 2022.

\bibitem{jonschkowski2015learning}
R.~Jonschkowski and O.~Brock, ``Learning state representations with robotic priors,'' \emph{Autonomous Robots}, vol.~39, pp. 407--428, 2015.

\bibitem{chen2020learning}
L.~Yen-Chen, A.~Zeng, S.~Song, P.~Isola, and T.-Y. Lin, ``Learning to see before learning to act: Visual pre-training for manipulation,'' in \emph{{IEEE International Conference on Robotics and Automation (ICRA)}}, 2020.

\bibitem{shah2021rrl}
R.~Shah and V.~Kumar, ``Rrl: Resnet as representation for reinforcement learning,'' in \emph{{Proceedings of International Conference on Machine Learning (ICML)}}, 2021.

\bibitem{parisi2022unsurprising}
S.~Parisi, A.~Rajeswaran, S.~Purushwalkam, and A.~Gupta, ``The unsurprising effectiveness of pre-trained vision models for control,'' in \emph{{Proceedings of International Conference on Machine Learning (ICML)}}, 2022.

\bibitem{seo2022reinforcement}
Y.~Seo, K.~Lee, S.~L. James, and P.~Abbeel, ``Reinforcement learning with action-free pre-training from videos,'' in \emph{{Proceedings of International Conference on Machine Learning (ICML)}}, 2022.

\bibitem{ViT}
A.~Dosovitskiy, L.~Beyer, A.~Kolesnikov, D.~Weissenborn, X.~Zhai, T.~Unterthiner, M.~Dehghani, M.~Minderer, G.~Heigold, S.~Gelly, J.~Uszkoreit, and N.~Houlsby, ``An image is worth 16x16 words: Transformers for image recognition at scale,'' in \emph{{Proceedings of International Conference on Learning Representations (ICLR)}}, 2021.

\bibitem{imagenet21k}
O.~Russakovsky, J.~Deng, H.~Su, J.~Krause, S.~Satheesh, S.~Ma, Z.~Huang, A.~Karpathy, A.~Khosla, M.~S. Bernstein, A.~C. Berg, and L.~Fei{-}Fei, ``Imagenet large scale visual recognition challenge,'' \emph{{International Journal of Computer Vision (IJCV)}}, vol. 115, no.~3, pp. 211--252, 2015.

\bibitem{lora}
E.~J. Hu, Y.~Shen, P.~Wallis, Z.~Allen{-}Zhu, Y.~Li, S.~Wang, L.~Wang, and W.~Chen, ``Lora: Low-rank adaptation of large language models,'' in \emph{{Proceedings of International Conference on Learning Representations (ICLR)}}, 2022.

\bibitem{epickitchen}
D.~Damen, H.~Doughty, G.~M. Farinella, S.~Fidler, A.~Furnari, E.~Kazakos, D.~Moltisanti, J.~Munro, T.~Perrett, W.~Price, and M.~Wray, ``The {EPIC-KITCHENS} dataset: Collection, challenges and baselines,'' \emph{{Transactions on Pattern Analysis and Machine Intelligence (TPAMI)}}, vol.~43, no.~11, pp. 4125--4141, 2021.

\bibitem{hamer}
G.~Pavlakos, D.~Shan, I.~Radosavovic, A.~Kanazawa, D.~Fouhey, and J.~Malik, ``Reconstructing hands in 3d with transformers,'' in \emph{{Proceedings of Conference on Computer Vision and Pattern Recognition (CVPR)}}, 2024.

\bibitem{yolov5}
Ultralytics, ``{YOLOv5},'' \url{https://github.com/ultralytics/yolov5}, 2020.

\bibitem{odise}
J.~Xu, S.~Liu, A.~Vahdat, W.~Byeon, X.~Wang, and S.~D. Mello, ``Open-vocabulary panoptic segmentation with text-to-image diffusion models,'' in \emph{{Proceedings of Conference on Computer Vision and Pattern Recognition (CVPR)}}, 2023.

\bibitem{e2fgvi}
Z.~Li, C.~Lu, J.~Qin, C.~Guo, and M.~Cheng, ``Towards an end-to-end framework for flow-guided video inpainting,'' in \emph{{Proceedings of Conference on Computer Vision and Pattern Recognition (CVPR)}}, 2022.

\bibitem{fang2020graspnet}
H.-S. Fang, C.~Wang, M.~Gou, and C.~Lu, ``Graspnet-1billion: A large-scale benchmark for general object grasping,'' in \emph{{Proceedings of Conference on Computer Vision and Pattern Recognition (CVPR)}}, 2020.

\bibitem{schulman2017proximal}
J.~Schulman, F.~Wolski, P.~Dhariwal, A.~Radford, and O.~Klimov, ``Proximal policy optimization algorithms,'' \emph{arXiv preprint arXiv:1707.06347}, 2017.

\bibitem{ma2023eureka}
Y.~J. Ma, W.~Liang, G.~Wang, D.~Huang, O.~Bastani, D.~Jayaraman, Y.~Zhu, L.~Fan, and A.~Anandkumar, ``Eureka: Human-level reward design via coding large language models,'' in \emph{{Proceedings of International Conference on Learning Representations (ICLR)}}, 2024.

\bibitem{spearman1961proof}
C.~Spearman, \emph{The proof and measurement of association between two things.}\hskip 1em plus 0.5em minus 0.4em\relax Appleton-Century-Crofts, 1961.

\end{thebibliography}

\end{document}